\documentclass[fleqn,10pt]{wlscirep}
\usepackage[utf8]{inputenc}
\usepackage[T1]{fontenc}
\usepackage{amsmath,amssymb,amsfonts}
\usepackage{graphicx}
\usepackage{booktabs}
\usepackage{multirow}
\usepackage{siunitx}
\usepackage{float}
\usepackage{xcolor}
\usepackage{hyperref}
\usepackage{url}

\title{LiDAR-based 3D Change Detection at City Scale}

\author[1,*]{Hezam Albaqami}
\author[2]{Haitian Wang}
\author[2]{Xinyu Wang}
\author[2]{Muhammad Ibrahim}
\author[3]{Zainy M. Malakan}
\author[1]{Abdullah M. Algamdi}
\author[4]{Mohammed H. Alghamdi}
\author[2]{Ajmal Mian}

\affil[1]{Department of Computer Science and Artificial Intelligence, College of Computer Science and Engineering, University of Jeddah, Jeddah 21493, Saudi Arabia}

\affil[2]{Department of Computer Science and Software Engineering, The University of Western Australia, Perth, WA 6009, Australia}
\affil[3]{Department of Data Science, Umm Al-Qura University, Makkah 24382, Saudi Arabia}
\affil[4]{Department of Information and Technology Systems, College of Computer Science and Engineering, University of Jeddah, Jeddah 21493, Saudi Arabia}

\affil[*]{haalbaqamii@uj.edu.sa}


\begin{abstract}
High-definition 3D city maps enable city planning and change detection which is essential for municipal compliance, map maintenance, and asset monitoring, including both built structures and urban greenery. Conventional Digital Surface Model (DSM) and image differencing are sensitive to vertical bias and viewpoint mismatch, while original point cloud or voxel models require large memory, assume perfect alignment, and degrade thin structures. We propose an uncertainty-aware, object-centric method for city-scale LiDAR-based change detection. Our method aligns data from different time periods using multi-resolution Normal Distributions Transform (NDT) and a point-to-plane Iterative Closest Point (ICP) method, normalizes elevation, and computes a per-point level of detection from registration covariance and surface roughness to calibrate change decisions.
Geometry-based associations are refined by semantic and instance segmentation and optimized using class-constrained bipartite assignment with augmented dummies to handle split–merge cases. Tiled processing bounds memory and preserves narrow ground changes, while instance-level decisions integrate overlap, displacement, and volumetric differences under local detection gating. We perform experiments on the city of Subiaco (Western Australia) dataset, captured once in 2023 and again in 2025. Our method achieves 95.3~\% accuracy, 90.8~\% mF1, and 82.9~\% mIoU, improving over the strongest baseline, Triplet KPConv, by 0.3, 0.6, and 1.1 points, respectively. The datasets are available on IEEE DataPort (2023 and 2025). The source code has been released on Github Repository.
\end{abstract}

\begin{document}

\flushbottom
\maketitle
\thispagestyle{empty}

\section*{Introduction}
\label{sec:introduction}

Modern cities are implementing smart transportation, digital twins, and autonomous driving systems that rely on high-definition (HD) 3D city maps maintained at an operational mapping cadence~\cite{DoTWA-AV-Strategy}. This creates the need for change detection i.e. 
identifying additions, removals, and volumetric variations across bitemporal 3D HD maps. Besides HD map maintenance, change detection is also useful for construction monitoring, municipal compliance, asset management, and timely updates for localization and planning in autonomy stacks~\cite{Zhang2021HDMapChange, DeGelis2021UrbanPCD, Janai2020AutonomousSurvey,SeifHu-Eng2016,Levinson-ICRA2007}. LiDAR (Light Detection And Ranging) sensor supports this process by providing dense 3D geometry--by driving a LiDAR mounted car through the city--that distinguishes long-term structural changes from transient dynamics such as vehicles and pedestrians~\cite{Ma2018MLSReview,Lague-ISPRS2013,Du-ISPRSJP2023}. 

\begin{figure}[!t]
    \centering \includegraphics[width=0.98\columnwidth,height=0.34\textheight,keepaspectratio]{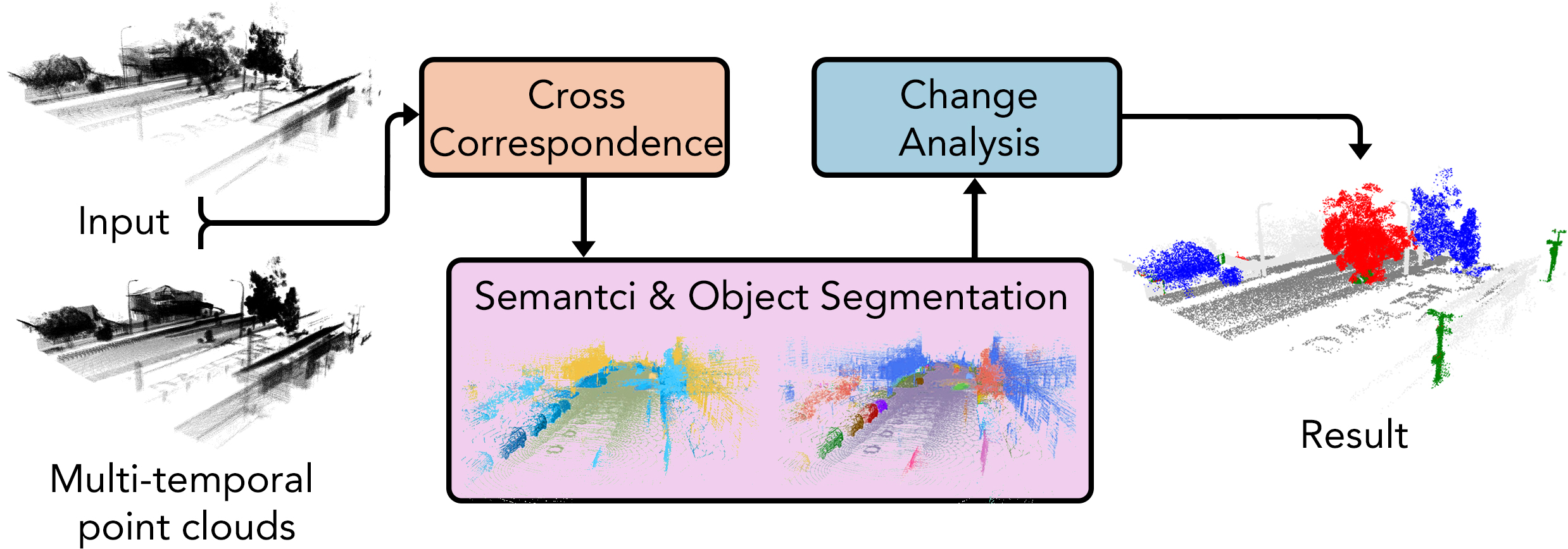}
    \vspace{-2mm}
    \caption{Overview of the proposed 3D change-detection pipeline. Analysing cross-epoch association, semantic and instance segmentation, and uncertainty-gated labels.}
    \label{fig:overview_2}
    \vspace{-2mm}
\end{figure}

Existing approaches to 3D change detection can be grouped into height- or DSM-based differencing, 2D image-based methods, and point-wise deep learning models~\cite{Xu2021PCDReview,Qin2016CDSurvey}. Height differencing demands precise co-registration under uniform sampling~\cite{Matikainen2010ALSBuildings,Murakami1999ALSChange}. In dense streets, small vertical biases and ground slopes cause systematic errors, while curbs and facade parallax yield false positives; moreover, results remain cell-wise without object identity~\cite{Lague2013M3C2,Matikainen2010ALSBuildings}. Image-based methods cannot resolve geometry and are vulnerable to viewpoint inconsistencies across satellites and time periods~\cite{Radke2005ImgCDSurvey,Daudt2018SiameseCD}. Point-wise Siamese and triplet networks, such as KPConv variants, process cropped tiles with millions of points~\cite{Thomas2019KPConv,Qi2017PointNet,Qi2017PointNetPP}. Their runtime and memory scale with area, training assumes balanced labels, and per-point predictions weaken Added/Removed reasoning when overlap is partial or occlusion alters sampling~\cite{Graham2018Submanifold,Choy2019Minkowski}. Voxel or bird’s-eye view encodings reduce memory but erode thin structures and fine-scale ground changes, forcing a trade-off between coverage and fidelity~\cite{Yan2018SECOND,Lang2019PointPillars}. 

Many pipelines also assume perfect pre-alignment and fail to propagate registration uncertainty into decision thresholds, leading to spurious detections near loop closures or under a canopy~\cite{Besl1992ICP,Chen1992Pt2Plane,Biber2003NDT,Segal2009GICP,Censi2007ICPcov}. Furthermore, few methods enforce class-consistent associations between epochs, leaving split and merge cases unresolved and causing per-class count drift at city scale~\cite{Teo2013FacadeChange,Weinmann2015SemanticSurvey,Xiao2015MLSChange3D}.  These constraints limit deployment for operational HD-map maintenance in dense urban corridors and highlight the need for a city-scale, object-centric, and uncertainty-aware approach.

We propose a change detection method that addresses the above limitations. Our method is object-centric and uncertainty-aware, operating at city scale on bi-temporal LiDAR maps. 

An overview of our method is shown in Fig.~\ref{fig:overview_2}.
The process begins with global alignment using multi-resolution NDT~\cite{zhu2011review} followed by point-to-plane ICP~\cite{segal2009generalized}. Both maps are then height-normalized, and a per-location level of detection is derived from registration covariance and surface roughness. This replaces the common assumption of perfect pre-alignment and introduces calibrated thresholds that suppress spurious detections along loops and under canopy. Geometry-only proxies initiate cross-epoch associations, followed by semantic and instance segmentation for ground, building, vegetation, and mobile classes on single-epoch geometry. Class-constrained bipartite matching refines correspondences and resolves split and merge cases through augmented assignment with dummies, thereby preventing per-class count drift and enforcing label consistency over time.

The city is partitioned into overlapping tiles of fixed size to bound memory and runtime, enabling dense street processing without erosion of thin structures that typically occurs with coarse voxel encodings. Change decisions are made per object using 3D occupancy overlap, normal displacement, and volumetric differences, each gated by the local level of detection. This object-level formulation resolves Added and Removed cases under partial overlap, mitigates slope- or curb-induced artifacts, and reduces false positives caused by viewpoint or sampling variations. The output is a per-instance change table and tiled maps labeled Added, Removed, Increased, Decreased, or Unchanged, each with confidence, aggregated city-wide to support HD-map maintenance and autonomous driving applications.

Another bottleneck in advancing outdoor city-scale change detection research is the lack of public benchmarks. Existing public benchmarks for 3D change detection cover limited area or rely on aerial data, and object-level annotations across data captured at different times remain scarce~\cite{DeGelis2023Urb3DCD, Xiao2015MLSChange3D}. Some of the datasets and methods work on simulated point clouds rather than detecting actual changes that occur over long periods of time \cite{Xiao2015MLSChange3D}. 
We also address this gap by releasing a follow-on Subiaco (Western Australia) city-scale dataset that extends our 2023 capture \cite{2hcq-5v45-25}. The new dataset was acquired in 2025, after a two-year interval. Subiaco, immediately west of Perth’s central business district, serves as a major transport hub with two arterial roads and heavy daily commuting, making it a representative urban testbed for long-term change analysis \cite{ciceklidag2024high}. 

We validate the proposed method on the bi-temporal city-scale maps of Subiaco from 2023 and 2025. A new 2025 HD LiDAR map is constructed and compared with the public 2023 counterpart~\cite{2hcq-5v45-25} under object-level annotations. Across 15 representative blocks, the method attains 95.3\% accuracy, 90.8\% macro F1, and 82.9\% macro IoU. When compared to the strongest baseline, Triplet KPConv, the improvements are 0.3 points in accuracy, 0.6 points in macro F1, and 1.1 points in macro IoU. The largest class-wise gain occurs for \textit{Decreased}, where the IoU reaches 75.6\%, exceeding the baseline by 8.4 points.

Our main contributions are as follows:
\begin{enumerate}
     \item \textbf{Method:} We propose an object-centric, uncertainty-aware 3D change detection method that integrates city-scale registration, local detection gating, class-consistent association, and instance-wise metrics on overlap, displacement, height, and volume. On 15 representative blocks from the Subiaco dataset, it achieves 95.3\% accuracy, 90.8\% mean F1, and 82.9\% mean IoU.
     
     \item \textbf{Dataset:} We propose a city-scale HD LiDAR dataset for city-scale change detection. Our new Subiaco 2025 dataset includes a globally referenced \texttt{.ply} map, per-instance semantics for \{ground, building, vegetation, mobile\}, and object-level change labels obtained by registering our new 2025 map to the public 2023 Subiaco map \cite{2hcq-5v45-25}. The 2025 Subiaco City Map datasets are available on \href{https://ieee-dataport.org/documents/2025-subiaco-wa-3d-hd-lidar-gnss-point-cloud-maps-dataset}{IEEE Dataport}. 
   
    \item \textbf{End-to-end Workflow:} We propose a reproducible end-to-end workflow for urban change mapping encompassing acquisition, map construction, semantic and instance segmentation, association, and change analysis with bounded-memory tiling. The source code has been released on \href{https://github.com/HaitianWang/IEEE-Sensor-Journal-Changing-Detection}{Github Repository}.
\end{enumerate}



\section*{Data Collection}
\label{sec:data_collection}

This section describes the acquisition design and data organization used to build the bi-temporal Subiaco dataset. We summarise the survey layout, multi-sensor platform, and logging protocol that support the construction of the 2025 reference map and the subsequent registration and object-level change analysis.

\begin{figure*}[t]
    \centering
    \includegraphics[width=\textwidth]{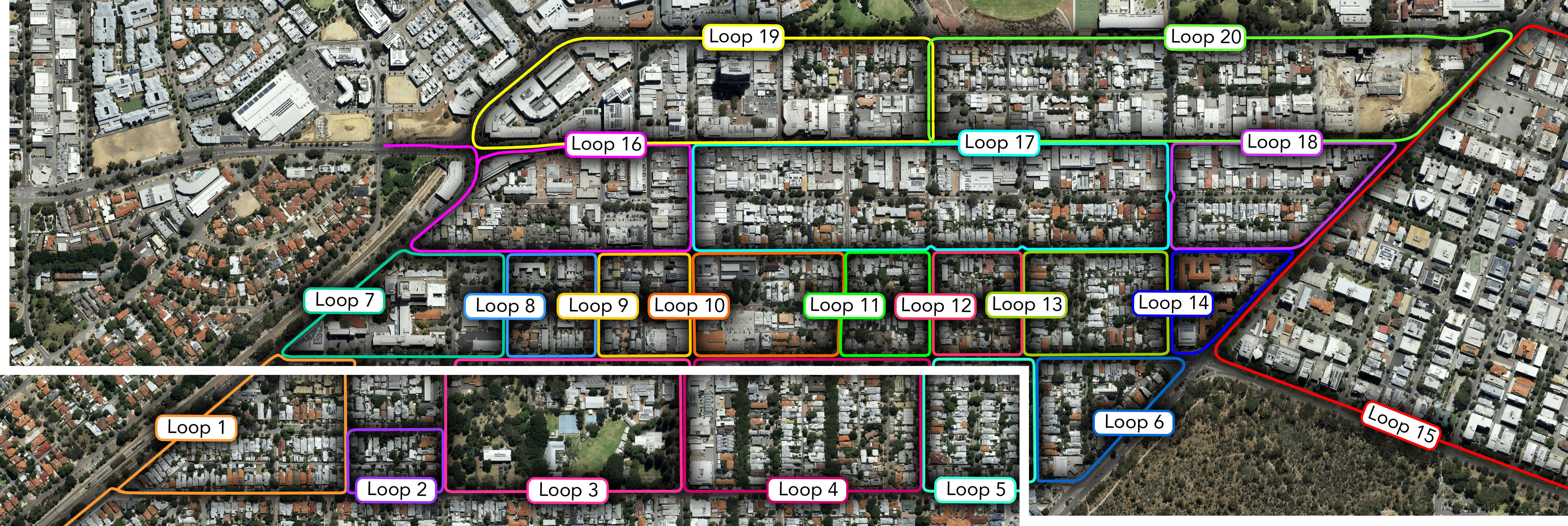}
    \vspace{-2mm}
    \caption{Subiaco driving loops for the 2025 data capture. Colored polylines denote twenty overlapping loops for individual reconstruction and merging. When comparing change detection results with other learning based methods, we take the mutually exclusive parts of Loop 1 to 5 for testing and the rest for training. }
    \label{fig:subi-routes}
    \vspace{-2mm}
\end{figure*}

\begin{figure}[t]
    \centering    \includegraphics[width=0.5\columnwidth,height=0.5\columnwidth,keepaspectratio=false]{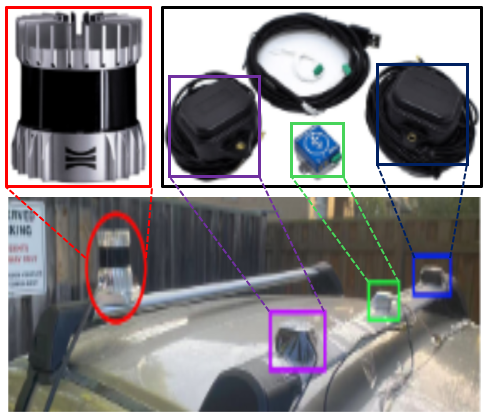}
    \vspace{-2mm}
    \caption{Multi-sensor rig. The system comprises an Ouster OS1-128 LiDAR, a dual-antenna RTK GNSS with MEMS IMU, and a logging unit mounted on a reinforced roof plate with a measured antenna baseline.}
    \vspace{-2mm}
    \label{fig:sensors-rig}
\end{figure}

\subsection{Survey Plan and Multi-Sensor Setup}

The survey covers Subiaco with overlapping road loops that span arterial corridors, residential blocks, and major intersections to force repeated viewpoints for loop closure and cross-street alignment. Fig~\ref{fig:subi-routes} illustrates the twenty loops used in 2025. Data were collected from a roof-mounted rig on a car driving at 10 to 30\,km/h during low-traffic windows. The primary sensor is an Ouster OS1-128 operated at 10\,Hz in dual-return mode with a $45^\circ$ vertical field of view, $0.35^\circ$ angular resolution, recording approximately 2.62\,M points per second. Geodetic reference and motion state come from a dual-antenna RTK GNSS with a MEMS IMU. GNSS fixes are logged at 5\,Hz and IMU at 73\,Hz. Sensors are rigidly mounted on a reinforced roof top rack with a measured antenna baseline to improve yaw observability, as shown in Fig~\ref{fig:sensors-rig}. Hardware time is synchronized with GNSS PPS. The LiDAR accepts the PPS and stamps packets with synchronized timestamps. All streams are recorded on an Ubuntu host using GigE for LiDAR and USB cable for GNSS and IMU. Factory extrinsics are refined by a short constrained-motion routine and verified by reprojection residuals on planar facades. Raw LiDAR \texttt{.pcap} packets are processed into per-scan \texttt{.ply} files containing dual returns and per-point timestamps. GNSS and IMU logs are exported to \texttt{.csv} format with UTC timestamps, quality indicators, and orientation measurements.

\vspace{-2mm}

\subsection{Data Acquisition Summary}
Each route is logged as an independent sequence with continuous 10\,Hz LiDAR frames, 5\,Hz GNSS fixes (RTK state recorded), and 73\,Hz IMU samples. LiDAR frames contain \mbox{260k–280k} points per scan with intensity and ring index. We retain both strongest and last returns; mapping uses strongest by default while both are archived for analysis. Sequences were recorded under dry daylight conditions to minimize rain and multipath artifacts; segments with RTK loss or packet drop exceeding pre-set thresholds are flagged in a YAML manifest. All trajectories and maps are georeferenced to GDA2020 / MGA Zone 50 and stored as globally referenced \texttt{.ply} maps together with per-sequence odometry, similar to the KITTI format \cite{geiger2013vision}. 

\begin{figure}[t]
    \centering \includegraphics[width=0.98\columnwidth,height=0.34\textheight,keepaspectratio]{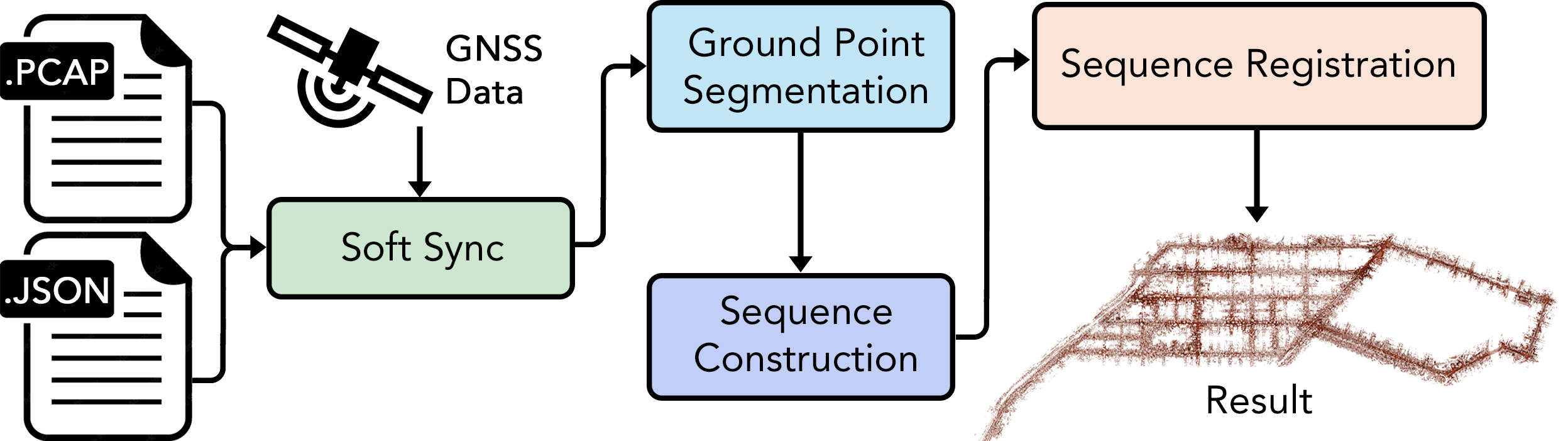}
    \vspace{-2mm}
    \caption{Pipeline of acquisition and mapping (soft GNSS synchronization, ground segmentation, sequence construction, NDT followed by ICP). Right:}
    \label{fig:overview_1}
    \vspace{-2mm}
\end{figure}

\vspace{-2mm}


\section*{Methods}
\label{sec:methods}

This section presents the proposed processing workflow after data acquisition. We first generate the 2025 city-scale reference map and then describe the geometric registration, semantic abstraction, class-consistent association, and uncertainty-gated change analysis used to produce the final instance-level decisions.

\subsection{3D City Map Generation}
Fig \ref{fig:overview_1} shows an overview of the map generation. LiDAR packets are deskewed using per-pixel timestamps and the closest IMU segment. Statistical outlier removal and a voxel pyramid (1.0/0.5/0.25\,m) standardize the density. Within each sequence, multi-resolution NDT~\cite{zhu2011review} provides stable coarse alignment; point-to-plane ICP~\cite{segal2009generalized} at each level. Loop closures are proposed by 360° LiDAR descriptors and verified by geometric consistency; a pose graph solves for globally consistent per-sequence poses. GNSS is used as a soft spatial prior: pose graph nodes that coincide with high-quality GNSS fixes (based on RTK status and dilution of precision) are anchored with adaptive weights, while GNSS-denied segments are down-weighted. Adjacent sequences are merged by aligning overlapping submaps with feature-initialized ICP and then fused. After global fusion, ground tiles are fit by robust plane models to define the vertical axis and zero height per tile; both epochs are rotated to align the vertical to $\mathbf{e}_z$ and are height-normalized to suppress long-wavelength bias. The final deliverable is a single globally referenced city-scale \texttt{.ply} map for 2025 at 0.25\,m leaf size with per-point coordinates, intensity, and normals, which is the input to registration/level of detection (LoD) estimation and the semantic layer.

\begin{figure*}[t]
  \centering
  \includegraphics[width=\textwidth,height=0.36\textheight,keepaspectratio]{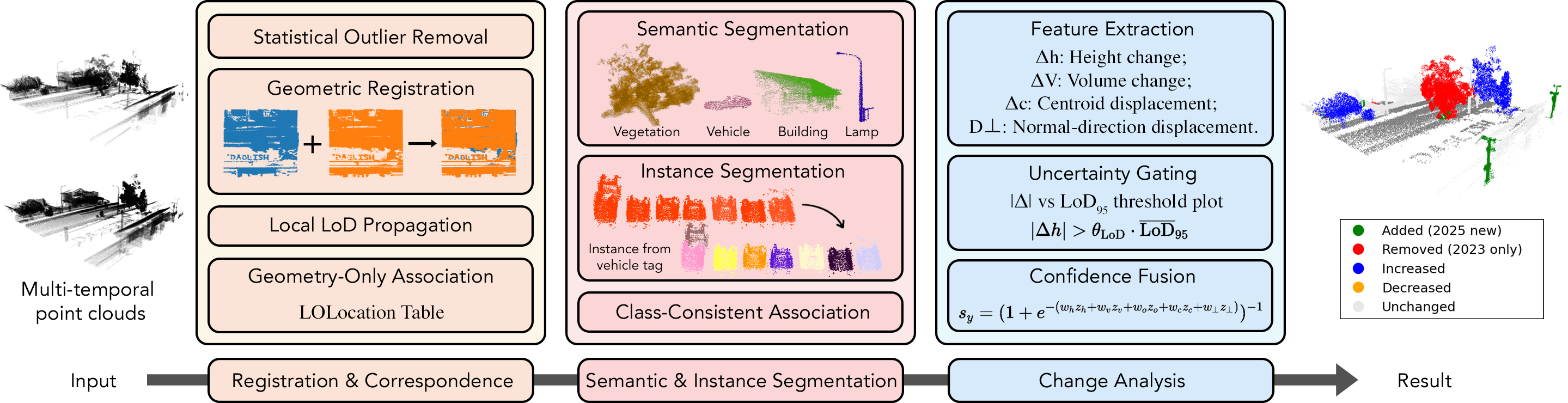}
  \vspace{-2mm}
  \caption{Object-centric, uncertainty-aware pipeline. Left-to-right: registration and correspondence, semantic/instance segmentation with class-consistent association, change analysis with LoD gating and confidence fusion, and the instance-level result.}
  \vspace{-2mm}
  \label{fig:pipeline}
\end{figure*}


\label{sec:methodology}
This section formalizes the proposed object centric change detection method on paired LiDAR maps. It introduces calibrated registration with a local level of detection to quantify uncertainty, derives geometry based correspondences, and constructs semantic and instance abstractions for ground, building, vegetation, and mobile classes. Class consistent association is then enforced and per object evidence from overlap, displacement, height, volume, and histogram cues is combined under uncertainty gating to produce five change labels with confidence. The design targets stability at city scale and supports reproducible evaluation and deployment. The proposed end-to-end workflow and data processing are summarized in Fig.~\ref{fig:pipeline}.

\subsection{Geometric Registration and Object-Level Correspondence}
\label{sec:data-layer}

Let $\mathcal{P}_{2023}$ and $\mathcal{P}_{2025}$ denote the city-scale LiDAR maps of Subiaco acquired in 2023 and 2025, respectively, each stored as a globally georeferenced \texttt{.ply}. This layer estimates a single rigid transform $T_{2023\rightarrow 2025}\!\in\! \mathrm{SE}(3)$ to align $\mathcal{P}_{2023}$ onto $\mathcal{P}_{2025}$, propagates registration uncertainty to a local level-of-detection, and derives level object location that will be refined by the semantic layer.

We first apply statistical outlier removal and build a voxel pyramid with leaf sizes $\{1.0,~0.50,~0.25\}$\,m to equalize density and reduce computation. Surface normals are estimated per level using a fixed-radius neighborhood ($r_n=0.8$\,m). Coarse alignment $T_0$ is obtained, when external priors are insufficient, by feature-based registration on the $1.0$\,m level: ISS keypoints, FPFH descriptors, and RANSAC with salient/support radii $\{1.5, 2.0\}$\,m and an inlier threshold of $0.6$\,m (20k iterations).

Precise alignment proceeds from coarse to fine with Normal Distributions Transform (NDT) followed by point-to-plane ICP. NDT supplies a stable pose update at each level; ICP then reduces the residual geometric error using target normals. Given correspondences $\{(p_i,q_i,n_i)\}$, we minimize
\begin{equation}
\min_{R\in SO(3),\,t\in\mathbb{R}^3}
\sum_{i} w_i\,\rho_{\tau}\!\big(n_i^\top (R\,p_i + t - q_i)\big),
\label{eq:pt2plane}
\end{equation}
where $w_i$ down-weights long-range pairs via a Huber kernel on Euclidean distance and $\rho_\tau(\cdot)$ is Tukey’s biweight with $\tau=0.3$\,m. We solve~\eqref{eq:pt2plane} by Gauss–Newton with analytic Jacobians and reject ill-conditioned updates via a normal-direction Fisher-information test. To stabilize long, low-curvature structures, a plane-to-plane penalty is added at the finest level after RANSAC plane extraction on facades and ground:
\begin{equation}
\mathcal{L}_{\text{pl2pl}}(R)=\lambda\sum_{k}\big\|R\,\bar{n}^{(2023)}_k-\bar{n}^{(2025)}_k\big\|_2^2,\qquad \lambda=0.1,
\end{equation}
and the total objective is $\mathcal{L}=\text{(\ref{eq:pt2plane})}+\mathcal{L}_{\text{pl2pl}}$. The final estimate is $T_{2023\rightarrow 2025}$ at $0.25$\,m. We report point-to-plane RMSE, inlier ratio, and the covariance-derived pose standard deviations.

After refinement, we enforce a consistent frame by robust ground-plane fitting on a $20{\times}20$\,m grid. The median ground normal defines the vertical axis; both epochs are rotated such that this axis aligns with $\mathbf{e}_z$. Heights are re-zeroed by subtracting the median ground elevation per tile to suppress long-wavelength vertical bias.

Registration uncertainty is propagated to a per-location LoD that will gate change claims in the analysis layer. For each evaluation site, the $95\%$ LoD along the local normal is
\begin{small}
\begin{align}
\mathrm{LoD}_{95}=1.96\sqrt{\frac{\sigma_{n,2023}^{2}}{N_{2023}}+\frac{\sigma_{n,2025}^{2}}{N_{2025}}+\sigma_{\mathrm{reg}}^{2}},
\label{eq:lod}
\end{align}
\end{small}
where $\sigma_{n,\cdot}$ denotes normal-direction roughness estimated in a cylindrical neighborhood (radius $1$\,m), $N_{\cdot}$ is the local sample size, and $\sigma_{\mathrm{reg}}$ is the global registration standard deviation from the final ICP Hessian. Equation~\eqref{eq:lod} is evaluated on a $5$\,m grid and bilinearly interpolated to points.

To initialize object-level associations prior to semantics, we extract geometry-only proxies. Ground cells are removed, and connected components are computed on a $0.5$\,m occupancy grid with morphological closing. For each component $o$ we compute a centroid $c(o)$, an Oriented Bounding Box (OBB) $\mathrm{OBB}(o)$ from Principal Component Analysis (PCA), the $95^{\text{th}}$ height quantile $h_{95}(o)$, and covariance eigen-features (linearity, planarity, sphericity). Cross-epoch association uses gated nearest neighbors with a shape-consistency cost
\begin{equation}
\label{eq:conf}
\begin{split}
s_y(o_i,o_j)
&=\Bigl(1+\exp\!\bigl[-\bigl(
w_h z_h + w_v z_v + w_o z_o + w_c z_c
{}+\, w_{\perp}\,\tfrac{|D_{\perp}(o_i,o_j)|}{\overline{\mathrm{LoD}}_{95}}
\bigr)\bigr]\Bigr)^{-1}.
\end{split}
\end{equation}
Here $c$ is the object centroid, $\mathrm{IoU}$ the 3D box overlap ratio, $H$ a 10-bin height histogram per object, and $D_{\chi^2}$ the chi-square distance between histograms. We set $(\alpha,\beta,\gamma)=(1,2,0.5)$ and accept a pair if $\|c_i-c_j\|_2<2$\,m, $\mathrm{IoU}>0.1$, and $\mathsf{cost}<3.5$. Unmatched components are treated as \textit{Added}/\textit{Removed}. Accepted pairs define $\pi:\mathcal{O}_{2023}\!\rightarrow\!\mathcal{O}_{2025}$ with attributes $(c,\mathrm{OBB},h_{95})$.

\subsection{Semantic and Instance Segmentation}
\label{sec:semantic-layer}

This layer assigns semantic classes (ground, building, vegetation, mobile) to points and produces instance-consistent objects per class. The design is unsupervised with geometry- and topology-based criteria; the annotated Subiaco maps are reserved for evaluation only. All decisions are made on single-epoch geometry and are later reconciled with cross-epoch correspondences before the change analysis.

We first compute per-point features on the finest voxel level (leaf $0.25$\,m): eigenvalues $(\lambda_1\!\ge\!\lambda_2\!\ge\!\lambda_3)$ of the local covariance (radius $0.6$\,m), normal $\mathbf{n}$, roughness $\sigma_n$ along $\mathbf{n}$, height above ground $z_g$ measured against the frame in Sec.~\ref{sec:data-layer}, and local density $\rho$. From $(\lambda_1,\lambda_2,\lambda_3)$ we derive linearity $L=(\lambda_1-\lambda_2)/\lambda_1$, planarity $P=(\lambda_2-\lambda_3)/\lambda_1$, and sphericity $S=\lambda_3/\lambda_1$. Points are grouped into superpoints by cut-pursuit on a $k$-NN graph ($k{=}20$) using an energy with a piecewise-constant data term on $(L,P,S,\|\mathbf{n}\cdot\mathbf{e}_z\|,z_g)$ and a Potts boundary term weighted by feature contrast. Let $\mathcal{V}$ be the node set (initial points) and $\mathcal{E}$ the edge set; the superpoint partition $\mathcal{U}$ minimizes
\begin{equation}
\label{eq:cp}
\begin{aligned}
\min_{\mathcal{U}} \quad
& \sum_{u\in\mathcal{U}} \sum_{i\in u} \left\lVert \mathbf{f}_i - \boldsymbol{\mu}_{u} \right\rVert_1
+ \eta \sum_{(i,j)\in\mathcal{E}} \omega_{ij}\,
\mathbb{1}\!\left[\mathcal{U}(i)\neq \mathcal{U}(j)\right],
\text{where} \quad
& \omega_{ij} = \exp\!\left(-\frac{\left\lVert \mathbf{f}_i-\mathbf{f}_j \right\rVert_2^2}{\sigma_f^2}\right).
\end{aligned}
\end{equation}
$\mathbf{f}_i=[L,P,S,\|\mathbf{n}_i\!\cdot\!\mathbf{e}_z\|,z_{g,i}]^\top$, $\boldsymbol{\mu}_u$ is the median feature of segment $u$, $\eta$ is the boundary weight, and $\sigma_f$ is set by median feature distance.

Semantic labeling is assigned to superpoints and then propagated to points. Ground is detected by robust plane support and slope: a superpoint is ground if its median distance to the local ground model is below $0.10$\,m and $\arccos(\|\mathbf{n}\cdot\mathbf{e}_z\|)>85^\circ$ on at least $80\%$ of its points; road markings and curbs adhere to ground via morphological closing in 2D tiles ($5{\times}5$\,m). Building is assigned to superpoints that satisfy $P>\tau_P$ with $\tau_P{=}0.6$, verticality $\|\mathbf{n}\cdot\mathbf{e}_z\|<0.25$, residual-to-plane $<0.08$\,m, and minimum planar area $>6$\,m$^2$ after merging coplanar neighbors by normal deviation $<10^\circ$. Vegetation is assigned to superpoints with $S>\tau_S$ where $\tau_S{=}0.35$, median $z_g>0.5$\,m, and high normal variance within the segment (95th–5th percentile of $\mathbf{n}$ azimuth difference $>30^\circ$). Remaining non-ground segments with compact size, moderate height, and weak planarity are set as mobile candidates if their oriented bounding box (OBB) satisfies: longest side $<5$\,m, height $<3$\,m, and volume $<60$\,m$^3$; segments with persistent strong planarity are reassigned to building. Class conflicts are resolved by a priority rule ${ground}\rightarrow{building}\rightarrow{vegetation}\rightarrow{mobile}$ applied only when thresholds are within $5\%$ of each other; otherwise the original assignment is kept.

\begin{figure}[!t]
  \centering
  \includegraphics[width=0.98\columnwidth,keepaspectratio]{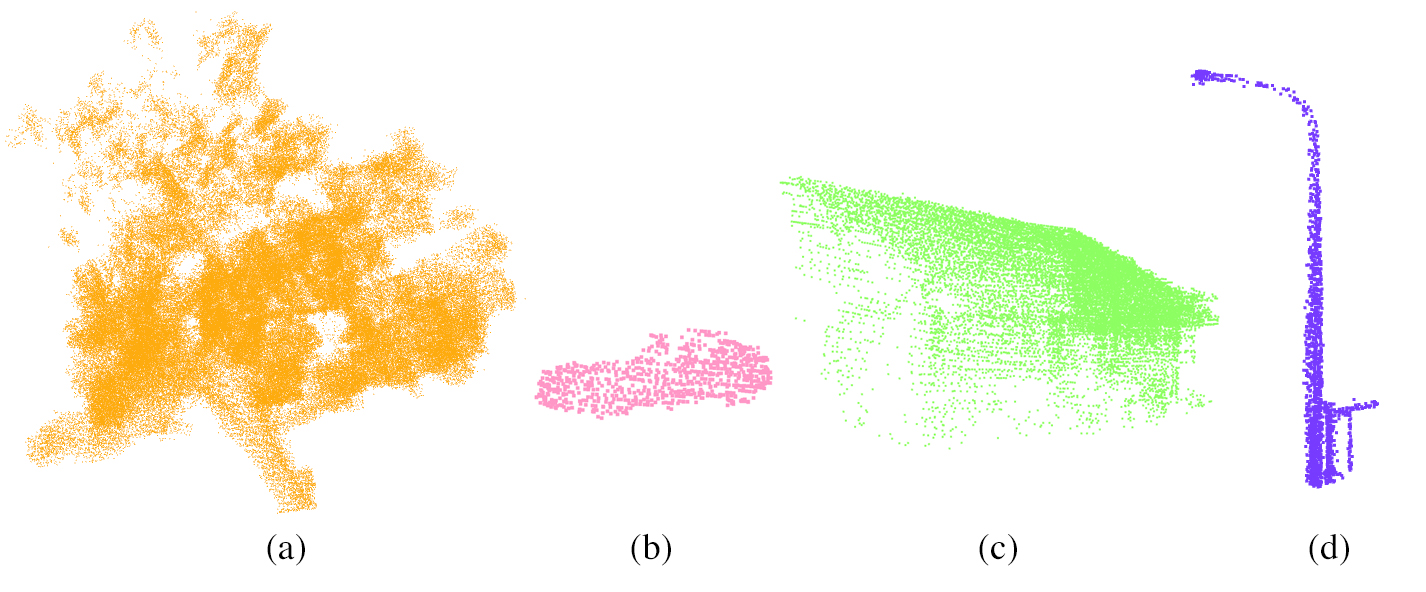}
  \vspace{-2mm}
  \caption{Examples of the four semantic categories: (a) vegetation, (b) mobile object, (c) building, and (d) other object (e.g., pole or ground fragment).}
  \label{fig:sem-examples}
  \vspace{-2mm}
\end{figure}

Fig~\ref{fig:sem-examples} illustrates typical outputs for the four semantic categories used in this work. Instances are produced per class by class-specific connectivity and geometry constraints. Ground instances are not required and remain a single support layer for later height and footprint queries. Building instances are formed by agglomerating coplanar facade patches through shared boundary length $>1.0$\,m and inter-patch normal deviation $<12^\circ$, then merging facade sets with overlapping horizontal OBB footprints (IoU$>0.3$) to form a building volume; small roof patches attached to the same footprint are merged if their normals are within $25^\circ$ of horizontal. Vegetation instances are extracted by Euclidean clustering with a distance threshold $d_c=\max(0.8\,\text{m},\,1.8/\sqrt[3]{\rho})$ and a minimum of $200$ points; touching crowns are separated by a marker-based opening in height slices of $1$\,m. Mobile instances are obtained with DBSCAN in $(x,y)$ with $\varepsilon=0.7$\,m and $\text{minPts}=30$, refined by removing attachments to the ground boundary via a $0.15$\,m erosion on the ground mask.

The level object location table is made class consistent by restricting associations to instances with the same semantic label and solving a regularized assignment problem per class. For class $c$, let $\mathcal{O}_{c,2023}$ and $\mathcal{O}_{c,2025}$ denote the instance sets at the two epochs. The binary matrix $\mathbf{X}\in\{0,1\}^{|\mathcal{O}_{c,2023}|\times|\mathcal{O}_{c,2025}|}$ encodes matches, and $u_i(\mathbf{X}) = 1-\sum_{j\in\mathcal{O}_{c,2025}} X_{ij}$ and $v_j(\mathbf{X}) = 1-\sum_{i\in\mathcal{O}_{c,2023}} X_{ij}$ measure unmatched instances on each side. The class-constrained association is obtained by
\begin{equation}
\label{eq:assign}
\begin{aligned}
\min_{\mathbf{X}} \quad &
\sum_{i\in\mathcal{O}_{c,2023}}\sum_{j\in\mathcal{O}_{c,2025}}
\big(\delta(o_i,o_j)+\phi(o_i,o_j)\big)X_{ij} 
&\;+\lambda\sum_{i\in\mathcal{O}_{c,2023}}u_i(\mathbf{X})
+\lambda\sum_{j\in\mathcal{O}_{c,2025}}v_j(\mathbf{X}) 
\text{s.t.}\quad &
\mathbf{X}\in\{0,1\}^{|\mathcal{O}_{c,2023}|\times|\mathcal{O}_{c,2025}|}.
\end{aligned}
\end{equation}

Here $\phi$ penalizes LoD-unsafe overlaps through
$\phi(o_i,o_j)=\max\big(0,\,\mathrm{LoD}_{95}(o_i,o_j)-d_{\perp}(o_i,o_j)\big)$, where $d_{\perp}(o_i,o_j)$ is the median normal-direction separation over the intersection of the two oriented bounding boxes. The weights are set to $\kappa=2.0$ and $\lambda=1.0$. Problem \eqref{eq:assign} is solved with the Hungarian algorithm on an augmented cost matrix with dummy rows and columns, and a match is accepted only if the 3D OBB IoU exceeds $0.05$ and the centroid distance is below $2$\,m. This produces per-instance, per-class correspondences that respect both semantics and geometry and are passed to the change analysis layer.

For all experiments on Subiaco the remaining parameters are fixed as follows. In \eqref{eq:cp} we use $(\eta,\sigma_f)=(0.8,\,\text{median}\,\|\mathbf{f}_i-\mathbf{f}_j\|_2)$. The ground plane residual is $0.10$\,m. The building plane residual is $0.08$\,m with minimum planar area $6$\,m$^2$. Vegetation uses $\tau_S=0.35$ and a crown split slice of $1$\,m. DBSCAN operates with $(\varepsilon,\text{minPts})=(0.7\,\text{m},30)$. The assignment parameters are $(\kappa,\lambda)=(2.0,1.0)$ with LoD gating given by \eqref{eq:lod}.

\subsection{Change Analysis}
\label{sec:change-analysis}

For each class $c\in\{{ground},{building},{vegetation},{mobile}\}$ and each associated pair $(o_{i,2023},o_{j,2025})$, we compute overlap, displacement, height, volume, and shape statistics at a fixed voxel size of $0.5$\,m and derive one of five labels $\{{Added},{Removed},{Increased},\\{Decreased},{Unchanged}\}$. The complete dataset is manually labeled into these five classes for evaluation of our method and comparison to other techniques. We also release these labels along with the Subiaco 2025 map dataset. All measurements are gated by the local level-of-detection $\mathrm{LoD}_{95}$ from Eq.~\eqref{eq:lod} to suppress pseudo-changes.

For overlapping geometry, we approximate the normal-direction displacement by sampling the intersection of the two oriented bounding boxes and projecting the local inter-epoch offset onto the average normal. Let $\Omega(o_i,o_j)$ be the set of voxel centers in the OBB intersection, $p_{2023}(x)$ and $p_{2025}(x)$ the nearest points to $x$ in $\mathcal{P}_{2023}$ and $\mathcal{P}_{2025}$, and $\bar{n}(x)$ the unit average of their estimated normals. The signed normal displacement is
\vspace{-1mm}
\begin{equation}
\vspace{-1mm}
D_{\perp}(o_i,o_j)=\operatorname*{median}_{x\in\Omega(o_i,o_j)}\;\bar{n}(x)^{\top}\!\left(p_{2025}(x)-p_{2023}(x)\right).
\label{eq:normal-disp}
\end{equation}
and we declare it informative only if $|D_{\perp}(o_i,o_j)|>\theta_{\text{LoD}}\cdot \overline{\mathrm{LoD}}_{95}(o_i,o_j)$ with $\theta_{\text{LoD}}=1.2$ and $\overline{\mathrm{LoD}}_{95}$ the median $\mathrm{LoD}_{95}$ over $\Omega(o_i,o_j)$. We compute 3D occupancy IoU on the union OBB at $0.5$\,m, centroid shift $\Delta \mathbf{c}=\mathbf{c}_j-\mathbf{c}_i$, height change $\Delta h=h^{2025}_{95}-h^{2023}_{95}$, volume change $\Delta V=V^{2025}-V^{2023}$ where $V$ is the occupied-voxel volume, and a chi-square distance between height histograms in the intersection OBB. All scalars are reported together with the per-class decision.

Decisions are class-specific and use fixed thresholds for Subiaco. For {building}, a pair is labeled {\em Increased} if $\mathrm{IoU}_{3\mathrm{D}}>0.10$ and either $|\Delta h|>0.50$\,m with $\Delta h>0$ and $|\Delta h|>\theta_{\text{LoD}}\cdot \overline{\mathrm{LoD}}_{95}$, or $\Delta V/V^{2023}>0.10$; it is labeled {\em Decreased} if $\mathrm{IoU}_{3\mathrm{D}}>0.10$ and $\Delta V/V^{2023}<-0.10$ or $\Delta h<-0.50$\,m with the same LoD gate. A pair is {\em Unchanged} if $\mathrm{IoU}_{3\mathrm{D}}>0.20$, $\|\Delta \mathbf{c}\|<1.5$\,m, $|\Delta h|\le 0.50$\,m, and $|\Delta V|/V^{2023}\le 0.10$, or if all informative statistics remain below their LoD. If $\mathrm{IoU}_{3\mathrm{D}}\le 0.10$ but a building instance exists only at 2025, it is {Added}; if it exists only at 2023, it is {Removed}. For {vegetation}, we keep the same logic with $(\mathrm{IoU}_{3\mathrm{D}}>0.08,\ \|\Delta \mathbf{c}\|<2.0$\,m$)$ and thresholds $(|\Delta h|>0.30$\,m, $|\Delta V|/V^{2023}>0.15)$; the histogram distance is used to break ties in favor of {Increased} when vertical mass shifts upward. For {ground}, changes are evaluated on a $2$\,m raster: a tile is {Increased} if the median ground elevation difference exceeds $+0.15$\,m over a contiguous area $>25$\,m$^2$; it is {Decreased} if the difference is below $-0.15$\,m with the same area constraint; otherwise {Unchanged}. Ground {Added}/{Removed} does not apply. For {mobile}, instances unmatched across epochs are labeled {Added}/{Removed}; matched instances are {Unchanged} if $\mathrm{IoU}_{3\mathrm{D}}>0.20$ and $\|\Delta \mathbf{c}\|<2.0$\,m, otherwise they are discarded from city-scale reporting since persistent mobiles are rare in multi-year intervals.

To aggregate heterogeneous evidence into a confidence score attached to each decision, we standardize the informative statistics by their uncertainty and combine them through a logistic map. Let $z_h=\frac{|\Delta h|}{\sqrt{\overline{\mathrm{LoD}}_{95}^{\,2}+\sigma_h^2}}$, $z_v=\frac{|\Delta V|}{\sqrt{(V^{2023}\cdot \sigma_v)^2+\epsilon}}$, $z_o=1-\mathrm{IoU}_{3\mathrm{D}}$, and $z_c=\frac{\max(0,\|\Delta \mathbf{c}\|-\tau_c)}{\tau_c}$ with $(\tau_c^{\text{bld}},\tau_c^{\text{veg}})=(1.5\,\text{m},2.0\,\text{m})$. The confidence for the selected label $y$ on pair $(o_i,o_j)$ is
\begin{equation}
\label{eq:conf}
\begin{split}
s_y(o_i,o_j)
= \Bigl(1+\exp\!\bigl[-\bigl(
w_h z_h + w_v z_v + w_o z_o + w_c z_c+\, w_{\perp}\,\tfrac{|D_{\perp}(o_i,o_j)|}{\overline{\mathrm{LoD}}_{95}}
\bigr)\bigr]\Bigr)^{-1}.
\end{split}
\end{equation}

with class-dependent weights fixed for Subiaco as $(w_h,w_v,w_o,w_c,w_{\perp})=(0.35,0.30,0.20,0.10,0.05)$ for buildings and $(0.30,0.35,0.15,0.10,0.10)$ for vegetation; for ground we use $(w_h,w_v,w_o,w_c,w_{\perp})=(0.50,0.00,0.25,0.00,0.25)$ with $\Delta h$ interpreted as median elevation difference on the tile. The score $s_y$ is reported jointly with the label.

The final outputs are a per-instance change table containing the instance identifiers, class, label, $\Delta h$, $\Delta V$, $\Delta \mathbf{c}$, $\mathrm{IoU}_{3\mathrm{D}}$, $D_{\perp}$, and $s_y$, and tiled maps for ground with per-tile elevation change and confidence. Instances that fail the LoD gate for all statistics are set to {Unchanged}.

\begin{figure*}[t]
\centering
\begin{minipage}[t]{0.495\textwidth}
  \centering
  \includegraphics[width=\linewidth,height=0.34\textheight,keepaspectratio]{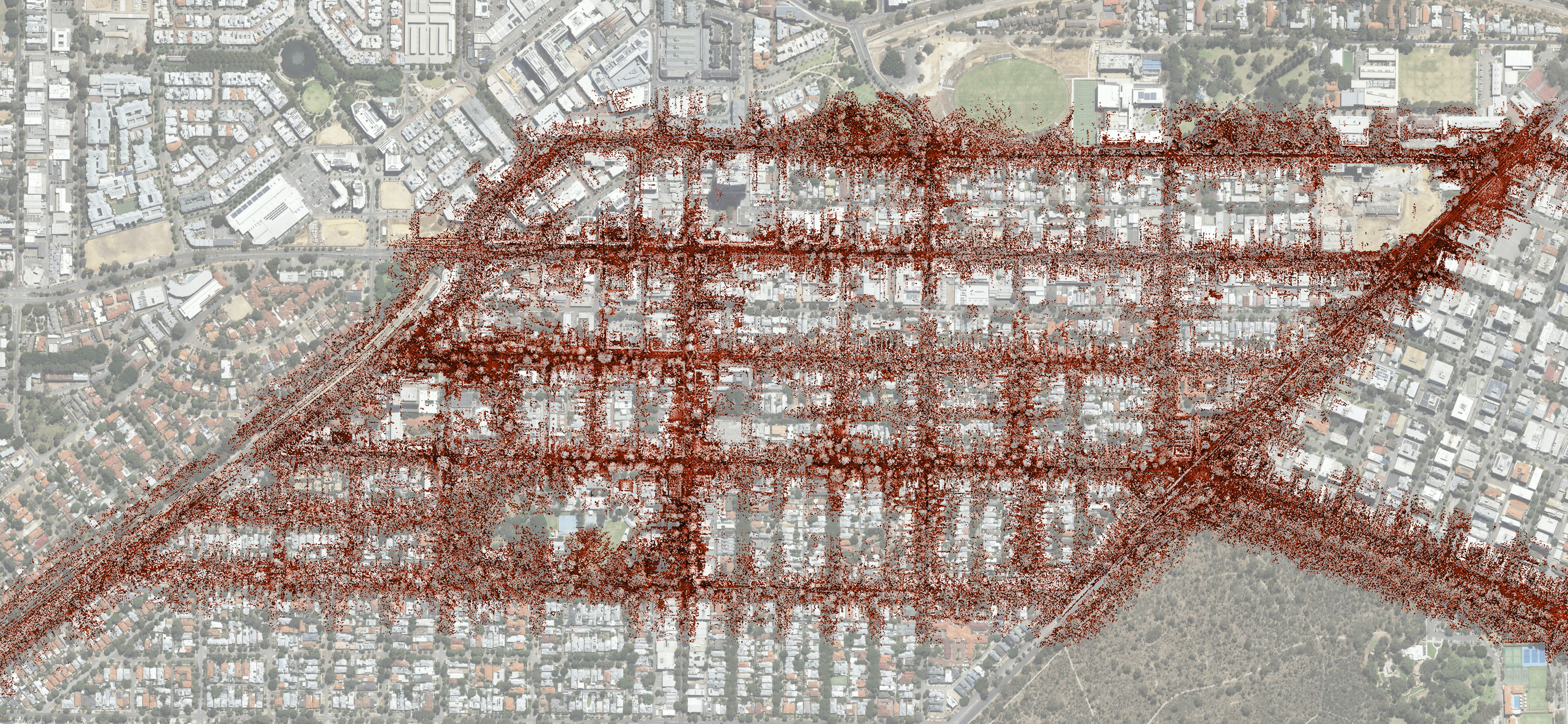}\\[-1mm]
  \footnotesize (a) 2025 fused and height–normalized 3D LiDAR-GNSS map~\cite{wwvk-0179-25}
\end{minipage}\hfill
\begin{minipage}[t]{0.495\textwidth}
  \centering
  \includegraphics[width=\linewidth,height=0.34\textheight,keepaspectratio]{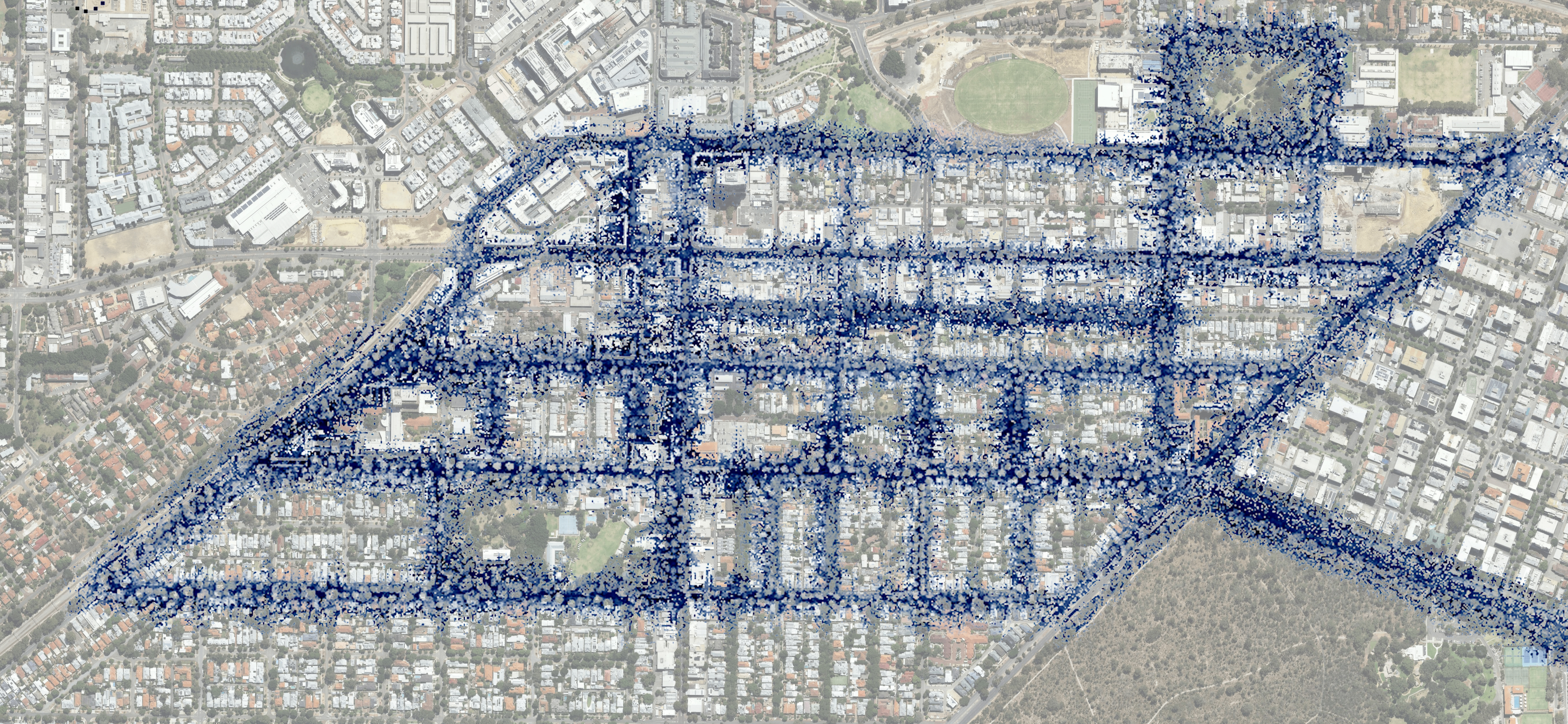}\\[-1mm]
  \footnotesize (b) 2023 3D LiDAR point cloud map~\cite{2hcq-5v45-25}
\end{minipage}
\vspace{-1mm}
\caption{City–scale reconstructions over Subiaco. The side-by-side layout avoids vertical overflows while keeping both panels within the page margins.}
\label{fig:subiaco-map-overview}
\vspace{-2mm}
\end{figure*}



\section*{Results}
\label{sec:results}
We evaluate our method on the proposed Subiaco map dataset and compare it to existing techniques using paired maps from 2023 and 2025 under a consistent protocol. Qualitative results on representative blocks illustrate typical urban patterns, and quantitative results report accuracy, macro F1, and macro IoU together with per class IoU. Comparisons with classical and point based baselines assess effectiveness across object categories and ground works, using fixed parameters and tiled processing to reflect the intended operational setting.

\subsection{Subiaco 2025 3D Map Construction Results}
The 2025 Subiaco map was fused into a single globally referenced point cloud at 0.25\,m voxel leaf size with per–point normals and intensity. LiDAR packets were deskewed using per–pixel timestamps and IMU. Outliers were removed and a 1.0/0.5/0.25\,m voxel pyramid standardized density. Each sequence was aligned by multi–resolution NDT followed by point–to–plane ICP. Loop closures from 360\textdegree{} LiDAR descriptors formed a pose graph. GNSS fixes with quality flags anchored the graph with adaptive weights. Overlapping submaps were aligned with feature–initialized ICP and fused. Ground tiles were fit by robust planes to define the vertical axis and to zero local heights, which eliminated long–wavelength vertical bias.

Fig~\ref{fig:subiaco-map-overview} compares the 2025 reconstruction with the 2023 map.  
The 2025 map shows continuous curb lines, coherent facades across contiguous blocks and closed intersections with consistent approach geometry. Roof ridges and small roof appendages are preserved. Under canopy the ground surface remains connected and tree crowns form complete outer envelopes. The 2023 overlay presents local shear near loop junctions, duplicated facade strips from drift and lateral offsets at intersections. The 2025 reconstruction is used as the geometric reference in all subsequent analyses.

\subsection{Qualitative Change Detection on Representative Block Samples}
To enable city–scale processing and visualization, both epochs are partitioned into 52 square blocks of \(80\times 80\)\,m with a \(10\)\,m overlap to avoid boundary artifacts during association and to keep each tile under memory limits at 0.25\,m voxels. All tiles are registered into the 2025 frame, and change labels are produced per object with the LoD gate and the class rules defined in the methodology. Results are visualized with five panels per block: raw projection, height map, semantic labels, instance labels, and the change map that encodes {\em Added} in green, {\em Removed} in red, {\em Increased} in blue, {\em Decreased} in orange, and {\em Unchanged} in gray. Block samples from loop 6, 7, and 8 are reported because they cover the typical patterns observed in Subiaco, namely building redevelopment around courtyards and street corners, vegetation growth and pruning under canopy, and roadway resurfacing at multi-lane intersections with parking bays. They also include strong occlusion, long facades, and tree rows, which are failure modes for point-wise differencing but are handled by object-level reasoning.

\begin{table}[t]
\centering
\caption{Performance on the Subiaco dataset across 20 loops. Metrics are accuracy (micro accuracy), mF1 (macro F1 across five change labels), and mIoU (macro IoU). Right-hand columns report per-class IoU. All values are percentages.}
\vspace{-2mm}
\label{tab:blocks_20}
\scriptsize
\setlength{\tabcolsep}{3.6pt}
\renewcommand{\arraystretch}{1.05}
\begin{tabular}{lcccccccc}
\toprule
\multirow{2}{*}{Loop} & \multirow{2}{*}{ACC} & \multirow{2}{*}{mF1} & \multirow{2}{*}{mIoU} & \multicolumn{5}{c}{Per Class IoU (\%)} \\
\cmidrule(lr){5-9}
 &  &  &  & Added & Removed & Increased & Decreased & Unchanged \\
\midrule
01 & 95.7 & 92.2 & 84.1 & 89.4 & 83.0 & 78.1 & 78.8 & 95.0 \\
02 & 96.0 & 92.0 & 83.8 & 89.1 & 82.7 & 77.8 & 78.1 & 96.9 \\
03 & 95.8 & 91.9 & 83.6 & 88.7 & 82.5 & 77.6 & 77.4 & 96.8 \\
04 & 96.2 & 92.3 & 84.3 & 89.6 & 83.2 & 78.4 & 79.2 & 95.1 \\
05 & 94.5 & 91.8 & 83.5 & 89.0 & 82.4 & 77.3 & 77.0 & 97.1 \\
06 & 95.1 & 90.3 & 82.4 & 88.4 & 81.8 & 75.3 & 74.0 & 92.5 \\
07 & 95.4 & 90.7 & 83.1 & 87.6 & 81.9 & 74.5 & 77.7 & 93.8 \\
08 & 94.8 & 89.9 & 81.6 & 86.7 & 81.8 & 74.0 & 70.2 & 95.3 \\
09 & 95.7 & 91.3 & 84.0 & 85.9 & 82.7 & 78.5 & 78.2 & 94.9 \\
10 & 94.1 & 88.6 & 79.6 & 87.2 & 82.7 & 75.9 & 60.3 & 92.1 \\
11 & 96.0 & 91.6 & 84.8 & 85.1 & 79.1 & 76.2 & 90.0 & 93.7 \\
12 & 94.5 & 89.4 & 80.7 & 85.9 & 79.1 & 76.3 & 69.2 & 92.9 \\
13 & 95.6 & 91.0 & 83.5 & 88.4 & 80.8 & 77.2 & 79.1 & 92.1 \\
14 & 95.2 & 90.8 & 83.0 & 88.4 & 78.6 & 75.7 & 76.3 & 96.0 \\
15 & 94.3 & 88.9 & 80.4 & 88.7 & 80.1 & 78.2 & 60.0 & 94.8 \\
16 & 94.9 & 90.1 & 81.9 & 87.4 & 82.4 & 78.2 & 68.5 & 93.2 \\
17 & 95.8 & 91.5 & 84.6 & 85.1 & 79.2 & 78.0 & 86.5 & 94.4 \\
18 & 95.3 & 90.7 & 82.9 & 87.2 & 81.5 & 77.4 & 75.7 & 92.7 \\
19 & 94.7 & 89.7 & 81.2 & 87.0 & 81.9 & 76.6 & 66.9 & 93.8 \\
20 & 95.9 & 91.4 & 84.8 & 85.1 & 78.2 & 77.5 & 89.4 & 94.0 \\
\midrule
\textbf{all} & \textbf{95.3} & \textbf{90.8} & \textbf{82.9} & \textbf{87.5} & \textbf{81.3} & \textbf{76.9} & \textbf{75.6} & \textbf{94.3} \\
\bottomrule
\end{tabular}
\vspace{-2mm}
\end{table}

\begin{figure*}[!t]
    \centering
    \includegraphics[width=0.9\textwidth]{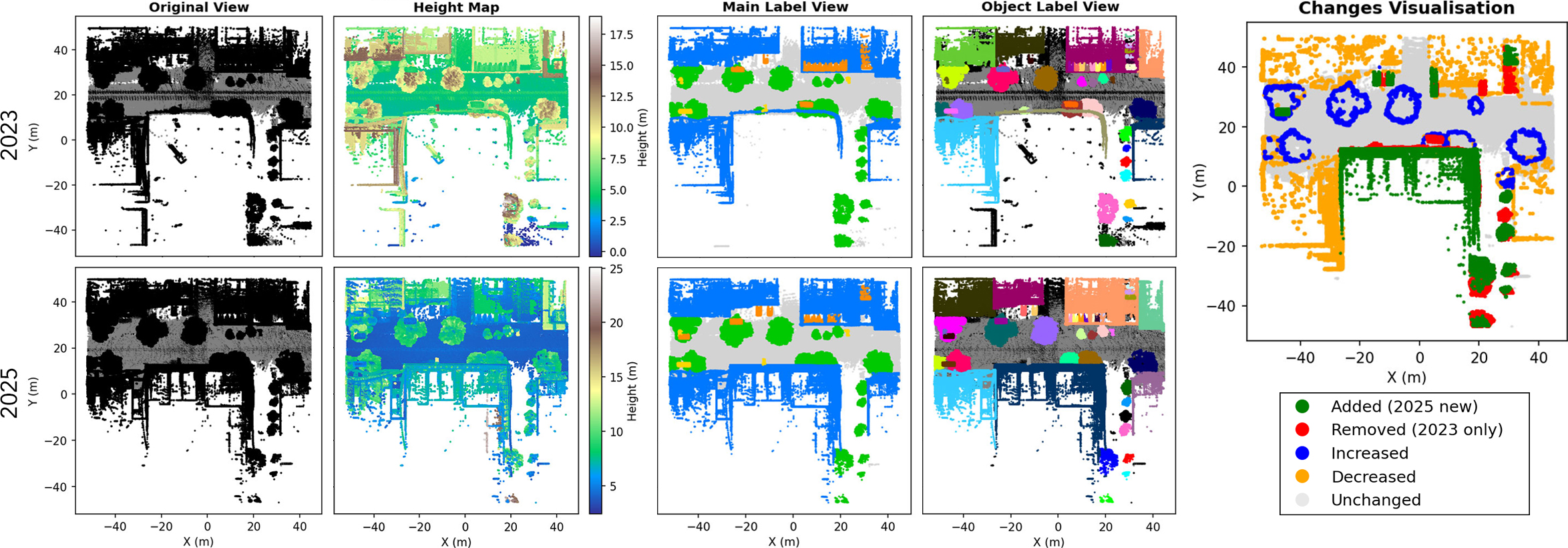}\vspace{2mm}
    \includegraphics[width=0.9\textwidth]{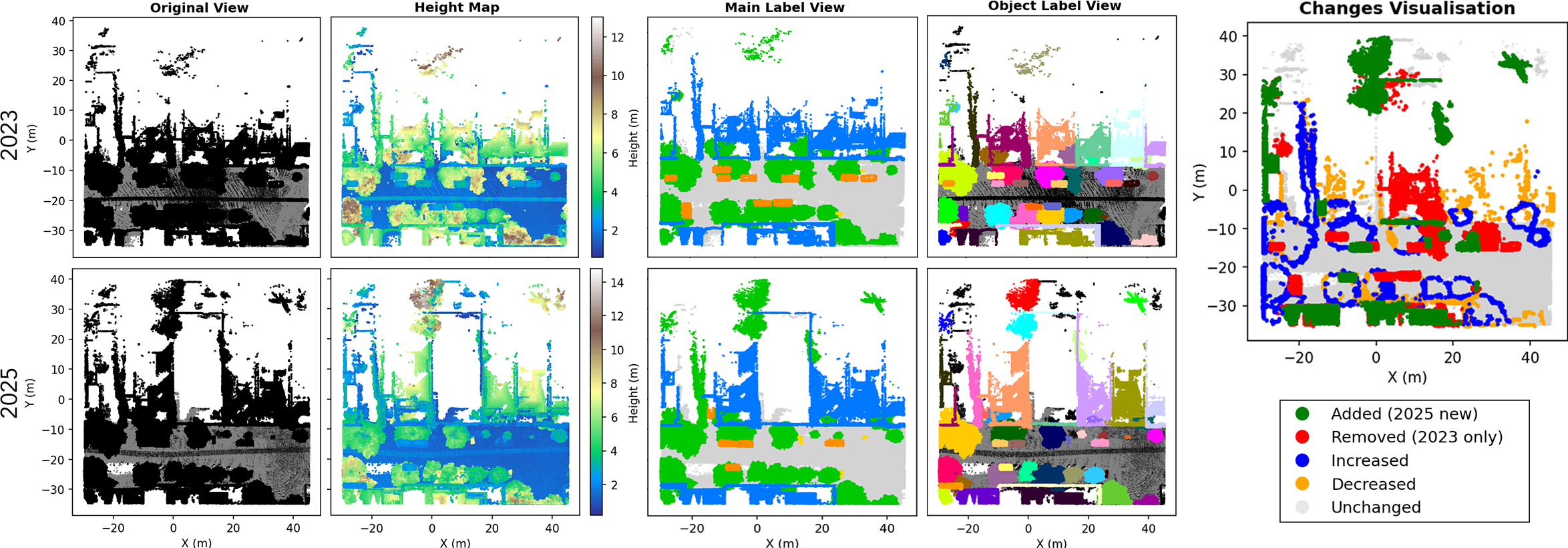}\vspace{2mm}
    \includegraphics[width=0.9\textwidth]{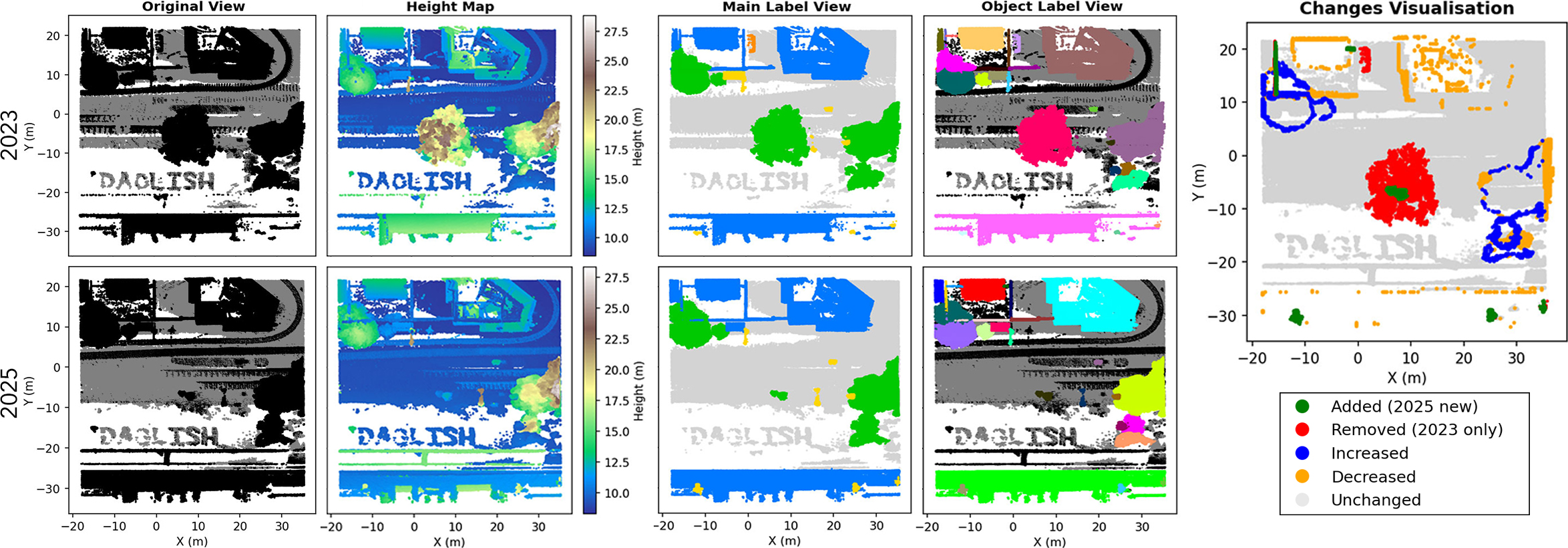}
    \vspace{-2mm}
    \caption{Qualitative change detection on three representative block samples. Each image presents the 2023 and 2025 projections, height maps, semantic and instance views, and the object-level change map with the legend embedded. From top to bottom: Samples from loop 6, 7, and 8.}
    \label{fig:blocks-qual}
    \vspace{-2mm}
\end{figure*}

Fig~\ref{fig:blocks-qual} shows qualitative outcomes for the three samples. The top one shows the method flags a large vegetation cluster as {\em Removed} at the center, multiple crowns along the verges as {Increased}, and narrow ground strips along the carriageway as {Decreased} consistent with resurfacing, while building volumes remain {\em Unchanged}. The middle visualization shows that new volumes within the inner lot are marked {\em Added} with height {\em Increased} at the northern frontage; shrubs along the southern footpath are {\em Removed}, the adjacent parking ground is {\em Decreased}, and surrounding facades retain high overlap. In the bottom section, the sample shows tree rows along the boulevard forming alternating {\em Added} and {Removed} crowns; the central median is reprofiled with contiguous {\em Increased} ground, and two small edge structures are {Removed}. Across the three tiles thin facade duplicates and boundary splits are suppressed by the overlap and by class-consistent assignment, and small ephemeral objects are ignored by the instance size and LoD thresholds. These examples illustrate the intended operating regime before the quantitative evaluation.

\begin{table*}[t]
\centering
\caption{Comparison with representative methods on the Subiaco dataset using identical metrics. All methods are evaluated on the first five loops (01--05). Accuracy is micro accuracy, mF1 is macro F1 across the five change labels, and mIoU is macro IoU. Right-hand columns report per-class IoU.}
\label{tab:method_comp}
\vspace{-2mm}
\small
\setlength{\tabcolsep}{5.2pt}     
\renewcommand{\arraystretch}{1.12}  

\begin{tabular*}{\textwidth}{@{\extracolsep{\fill}} 
l l
S[table-format=2.1]
S[table-format=2.1]
S[table-format=2.1]
*{5}{S[table-format=2.1]}
}
\toprule
Method & Source & \multicolumn{1}{c}{ACC} & \multicolumn{1}{c}{mF1} & \multicolumn{1}{c}{mIoU}
& \multicolumn{5}{c}{Per-Class IoU (\%)} \\
\cmidrule(lr){6-10}
 &  &  &  &  & {Added} & {Removed} & {Increased} & {Decreased} & {Unchanged} \\
\midrule
DSM--Siamese~\cite{Daudt2018SiameseCD} & ICIP'18  & 92.1 & 73.9 & 58.3 & 69.4 & 54.2 & 29.7 & 46.1 & 92.1 \\
DSM--FC--EF~\cite{Daudt2019FCEF} & TGRS'19 & 92.2 & 72.7 & 57.2 & 68.7 & 52.3 & 27.9 & 43.5 & 93.6 \\
Siamese KPConv~\cite{Thomas2019KPConv} & ICCV'19 & 94.6 & 89.0 & 80.1 & 85.9 & 78.4 & 72.8 & 68.1 & 95.5 \\
Triplet KPConv~\cite{Wu2021TripletKPConv} & ISPRS'21 & 95.0 & 90.2 & 81.8 & 88.4 & 82.1 & 75.6 & 67.2 & 95.9 \\
DC3DCD EFSKPConv~\cite{Kharroubi2025DC3DCD} & ISPRS'25 & 92.0 & 74.1 & 58.9 & 69.1 & 53.2 & 31.0 & 47.2 & 93.9 \\
\midrule
\textbf{Object--based (ours)} & --- & \textbf{95.6} & \textbf{92.0} & \textbf{83.9} & \textbf{89.1} & \textbf{82.8} & \textbf{77.8} & \textbf{78.1} & \textbf{96.2} \\
\bottomrule
\end{tabular*}

\vspace{-2mm}
\end{table*}

\subsection{Quantitative Evaluation on all 21 Loops}

We first evaluate the proposed method on all twenty loops using the five change labels. Accuracy, macro F1 and macro IoU are computed per loop, and per class IoU is reported to expose systematic errors. Table~\ref{tab:blocks_20} shows that accuracy remains within 94.1\% to 96.2\%, macro F1 within 88.6\% to 92.3\%, and macro IoU within 79.6\% to 84.8\%. The {\em Unchanged} class consistently exceeds 92\% IoU, indicating stable behavior on non events. Lower scores on {\em Decreased} reflect narrow resurfacing and partial curb overlap where geometry is sparse. Aggregating all loops yields 95.3\% accuracy, 90.8\% macro F1 and 82.9\% macro IoU, with class IoUs of 87.5\% for {\em Added}, 81.3\% for {\em Removed}, 76.9\% for {\em Increased}, 75.6\% for {\em Decreased} and 94.3\% for {\em Unchanged}. These results quantify the performance of our model when trained on loops 6 to 20 and evaluated across the full set.
        
\subsection{Comparison to Learning based Methods on the Test Set}
Although our method does not require training data, for comparison to learning based methods, we partitioned the map into mutually exclusive training and test sets as shown in Fig \ref{fig:subi-routes}. In Table~\ref{tab:method_comp}, we compare our method to learning based methods on the test partition. Classical DSM based and RF based approaches remain below 60\% macro IoU. KPConv and its variants increase performance to the low 80s but remain sensitive to thin ground changes. Our object based method reaches 95.6\% accuracy, 92.0\% macro F1 and 83.9\% macro IoU on the same test set. Relative to Triplet KPConv, the gains are 0.6 points in accuracy, 1.8 points in macro F1 and 2.1 points in macro IoU. The largest improvement appears on the {\em Decreased} category where our method achieves 78.1\% IoU, outperforming the nearest competitor by 10\% points. Our method reduces false detections on narrow resurfaced strips while maintaining consistent results across the remaining classes.


\section*{Discussion}
\label{sec:discussion}
This paper presented a method for 3D change detection using an object level, uncertainty aware approach on bi-temporal LiDAR maps at city scale. Our approach replaces perfect pre-alignment assumptions with calibrated registration and a local level of detection. This enforces class consistent instance association to resolve split and merge cases. Our method applies tiled processing with per object overlap, displacement, height, and volume cues to bound memory while preserving narrow ground changes. We also proposed a new benchmark dataset to fill a significant gap in outdoor urban change detection research. Our method outperformed existing techniques on this challenging benchmark, highlighting the promise of our end-to-end workflow. The new Subiaco 2025 dataset and our implementation will enable reproducible evaluation and support HD map maintenance in addition to change detection. Future work will extend to multi epoch sequences for trend analysis, fuse aerial and ground sensing, and learn to incorporate spatially varying uncertainty for online map updates.



\section*{Author contributions statement}
H.A. conceived the study. H.A., H.W., X.W., and M.I. developed the methodology and conducted the experiments. H.W. and X.W. prepared the dataset and implemented the core pipeline. Z.M.M., A.M.A., M.H.A., and A.M. contributed to analysis, interpretation, and supervision. All authors reviewed and approved the manuscript.

\section*{Additional information}
\textbf{Competing interests:} The authors declare no competing interests.

\section*{Funding:}
This work was funded by the University of Jeddah, Jeddah, Saudi Arabia, under grant No.
(UJ-24-SUTU-1290). The authors, therefore, thank the University of Jeddah for its technical
and financial support. This project is a collaborative effort between the University of Jeddah and the University of Western Australia, conducted under the Sustainable Partnerships Program with Distinguished International Universities.

\bibliography{main}

\end{document}